\RequirePackage[T1]{fontenc}
\documentclass[lettersize,journal]{IEEEtran}
\usepackage{amsmath,amsfonts}
\usepackage{array}
\usepackage[caption=false,font=normalsize,labelfont=sf,textfont=sf]{subfig}
\usepackage{textcomp}
\usepackage{stfloats}
\usepackage{url}
\usepackage{verbatim}
\usepackage{graphicx}
\usepackage{cite}
\usepackage{amsfonts}
\usepackage{amsmath}
\usepackage{amssymb}
\usepackage{ragged2e}
\usepackage{booktabs,makecell, multirow, tabularx}
\usepackage{float}
\usepackage{adjustbox}
\usepackage{epstopdf}
\usepackage{bibentry}
\usepackage{graphicx}
\usepackage{colortbl} %
\usepackage{xcolor}   %
\usepackage{amsmath, amsthm}
\usepackage[ruled,vlined]{algorithm2e}
\usepackage{algpseudocode}
\newtheorem{theorem}{Theorem}

\newtheorem{proposition}{Proposition}

\usepackage{newunicodechar}
\newunicodechar{–}{--}
\newunicodechar{’}{'}
\begin{document}

\title{RSWP: Robust SO(3)-Equivariant Watermarking for Panoramic Images}

\author{Pengzhen~Chen,
        Yanwei~Liu,
        Xiaoyan~Gu,
        Jinxia~Liu,
        Siwei~Ma,~\IEEEmembership{Fellow,~IEEE,}
        Weiping~Wang,
        and~Wu~Liu,~\IEEEmembership{Senior Member,~IEEE}%

\thanks{Pengzhen~Chen, Yanwei~Liu, Xiaoyan~Gu and Weiping~Wang are with the Institute of Information Engineering, Chinese Academy of Sciences; the School of Cyber Security, University of the Chinese Academy of Sciences; State Key Laboratory of Cyberspace Security Defense, Beijing 100093, China (e-mail: chenpengzhen@iie.ac.cn; liuyanwei@iie.ac.cn; guxiaoyan@iie.ac.cn; wangweiping@iie.ac.cn).}

\thanks{Jinxia~Liu is with the College of Information and Intelligence Engineering, Zhejiang Wanli University, Ningbo 315100, China (e-mail: liujinxia1969@126.com).}%
\thanks{Siwei~Ma is with the National Engineering Research Center of Visual Technology, Peking University, Beijing 100871, China (e-mail: swma@pku.edu.cn).}%
\thanks{Wu Liu is with the School of Information Science and Technology, University of Science and Technology of China, Hefei 230022, China (e-mail: liuwu@ustc.edu.cn).}
}%

\maketitle
\begin{abstract}

Robust watermarking for panoramic imagery is impeded by the topological mismatch between spherical signals and planar representations. Conventional methods fail under $SO(3)$ viewing rotations due to non-linear geometric warping and struggle with viewport extraction, which typically necessitates computationally expensive inverse remapping. To address these challenges, we propose RSWP, a watermarking framework grounded in icosahedral group theory. Leveraging the property that the icosahedral group ($\mathcal{I}_{60}$) serves as the largest discrete subgroup of $SO(3)$, we decompose spherical rotations into tractable local planar rotations and combinatorial facet permutations. Accordingly, we introduce a decoupling architecture that embeds watermarks into discrete facets and a theoretically guaranteed Equivariance-to-Invariance decoding paradigm. By employing an equivariant feature extractor followed by a permutation-invariant sorting module, this paradigm rigorously synchronizes rotation-distorted features. Moreover, the dual-extractor design inherently supports direct, blind watermark retrieval from partial 2D renderings without geometric calibration. Extensive experiments demonstrate that RSWP achieves state-of-the-art robustness against continuous $SO(3)$ rotations and viewport transformations while preserving high visual fidelity.

\end{abstract}


\section{Introduction}

The rapid proliferation of Virtual Reality (VR) systems and the emergence of the Metaverse paradigm~\cite{vr} have fundamentally transformed the creation, distribution, and consumption of visual content. Visual representations are transitioning from conventional planar images with limited fields of view (FoV) to immersive $360^{\circ}$ panoramic imagery, which captures complete surrounding environments. While originally driven by applications such as virtual tours and immersive entertainment, panoramic imagery has since evolved into a critical visual modality underpinning a wide spectrum of real-world and artificial intelligence systems.

Specifically, it is playing an indispensable role across diverse domains, including autonomous driving perception~\cite{ad1}, industrial inspection, medical endoscopy, and satellite-based environmental monitoring~\cite{app1}. More importantly, omnidirectional data has become a foundational substrate for  Artificial General Intelligence (AGI). In particular, panoramic observations provide globally consistent spatial context and long-term spatiotemporal information, which are essential for learning world models, also enabling robust Simultaneous Localization and Mapping (SLAM), and supporting Vision-and-Language Navigation (VLN) in embodied AI agents operating in complex, unstructured environments~\cite{AiCW25,yao2025navmorph,gen}.

This transition has been further accelerated by the advent of generative techniques. Recent diffusion-based architectures, such as DiffPano~\cite{ye2024diffpano} and PanFusion~\cite{zhang2024taming}, utilize spherical epipolar constraints to synthesize high-fidelity, globally consistent panoramic environments from textual descriptions. These advancements have drastically reduced the cost of 3D asset creation, fueling downstream applications ranging from digital twin construction to autonomous driving simulation \cite{ad1}. Consequently, high-quality panoramic datasets have evolved from mere visual assets to high-value spatial intelligence training corpora, serving as the bridge between digital simulation and physical reality.

However, the democratization of high-quality panoramic generation and the strategic importance of these datasets have precipitated a severe crisis regarding copyright and data ownership. The ability to synthesize photorealistic environments facilitates the unauthorized utilization of proprietary real-world scans. Malicious actors can scrape copyrighted panoramic assets to train rival World Models, Neural Radiance Fields (NeRF) and 3D Gaussian Splatting (3DGS) models without consent, effectively ``laundering" intellectual property through neural network weights \cite{jang20253d}. Therefore, establishing robust mechanisms for copyright protection and ownership verification in the spherical domain has become an urgent imperative to safeguard the integrity of this critical 3D infrastructure.

\begin{figure*}[t] %

	\centering %
	\includegraphics[width=\textwidth]{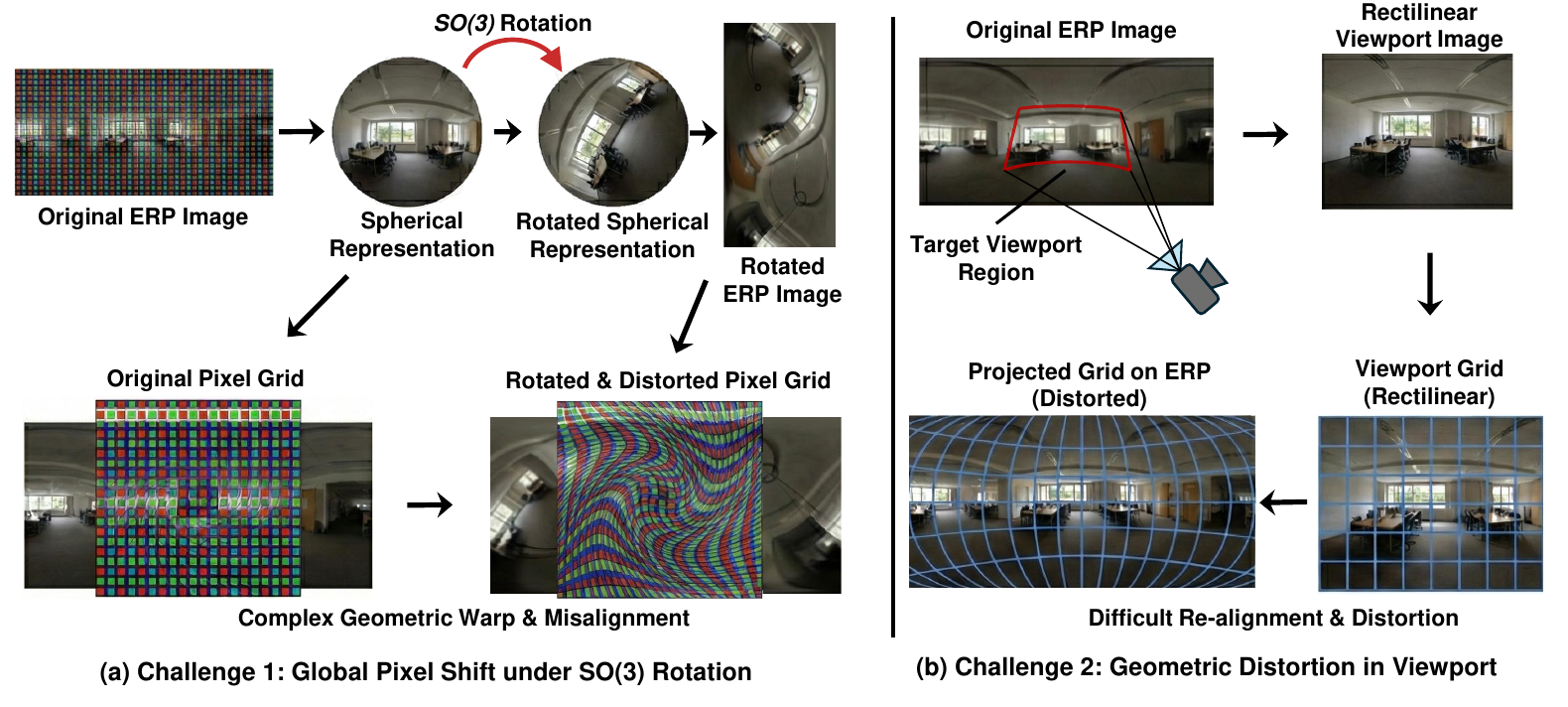} %
	\caption{Inherent geometric challenges in panoramic watermarking. (a) Global $SO(3)$ Rotation: A rigid rotation on the spherical domain manifests as complex, non-linear warping and pixel misalignment when projected back to the 2D ERP format. (b) Viewport Rendering: The projection from the curvilinear ERP domain to a rectilinear viewport introduces severe geometric discrepancies, distorting the spatial grid structure required for standard watermark extraction.} %
	\label{fig:rotate} %
\end{figure*}

Digital watermarking is a standard defense mechanism for copyright protection. While existing watermarking algorithms~\cite{luo2026diffw,wm3,mbrs} exhibit robustness against common attacks such as compression and Gaussian noise in conventional 2D planar images, they are fundamentally inadequate for panoramic imagery due to intrinsic geometric and topological differences.
A panoramic image represents a signal defined on a sphere ($S^2$), which is typically projected onto a 2D plane via Equirectangular Projection (ERP) for storage and transmission. In applicable scenarios, users can freely alter their viewing direction via head-mounted displays, which leads to two domain-inherent transformations: global $SO(3)$ rotations and partial viewport rendering. First,  user's interactions with panoramic content typically involve looking around, which corresponds to rotating the signal on the sphere ($SO(3)$ group~\cite{eq6}). When represented in ERP format, however, such rigid rotations on the sphere manifest as complex, non-linear geometric warping and translational misalignment as shown in Fig.~\ref{fig:rotate}(a). Unlike simple 2D affine transformations, these distortions vary spatially across the image, causing embedded watermark patterns to undergo non-uniform stretching and twisting.
Consequently, the pixel-grid synchronization required by conventional decoders is destroyed, rendering CNN-based watermarking
methods, whose robustness relies primarily on translational equivariance, ineffective under any-angle rotations.
Second, practical VR systems display only localized fields of view, referred to as viewports~\cite{viewport}. As illustrated in Fig.~\ref{fig:rotate}(b), a viewport introduces substantial geometric discrepancies compared to the corresponding ERP region. Watermarks embedded in the curvilinear ERP domain appear geometrically deformed when projected onto the rectilinear viewport plane. This domain-mismatch breaks the structural integrity of the watermark signal, rendering traditional 2D watermarking schemes ineffective. Existing panoramic watermarking approaches~\cite{pa1,pa2} attempt to mitigate this issue by remapping viewports back to the spherical or ERP domain prior to extraction. However, such strategies require precise geometric calibration and access to the original panoramic layout, violating the requirements of blind watermarking and significantly limiting robustness in real-time applications.

To address these challenges, we propose RSWP, a Robust SO(3)-Equivariant Watermarking Framework for Panoramic Images. Our approach decomposes panoramic images into an icosahedral projection-based representation and embeds watermarks into its planar facets. By exploiting the symmetry of the icosahedral group, we design an SO(3)-equivariant decoding network to extract watermarks robust against SO(3) rotations. Additionally, the locally planar nature of icosahedral facets provides a close geometric approximation to both the spherical surface and rectilinear viewports, allowing watermarks to be directly extracted from viewport images without remapping to the global panoramic domain. This design fundamentally overcomes the limitations of existing methods and establishes a geometry-aware watermarking paradigm for omnidirectional imagery. In summary, our key contributions are as follows:

\begin{itemize}
\item We propose RSWP, the first $SO(3)$-equivariant watermarking framework for panoramic images that achieves robustness to large-angle $SO(3)$ rotations and calibration-free watermark retrieval directly from viewports.

\item We introduce a novel Equivariance-to-Invariance decoding paradigm. Grounded in icosahedral group theory, it mathematically overcomes the spherical-planar topological mismatch by decoupling 3D rotations into local planar rotations and combinatorial facet permutations.

\item  Extensive experiments demonstrate that RSWP achieves superior performance in visual fidelity, watermark extraction accuracy, and robustness against both $SO(3)$ rotations and viewport transformations.

\end{itemize}

\section{Related Work}

\subsection{Watermarking for Planar Images and 3D Content}
Digital watermarking spans methods based on geometric moment invariants~\cite{he2024invariants} and learned embedding and extraction frameworks~\cite{wm3,luo2026diffw}, with recent work addressing copyright protection and robustness under image editing and generation~\cite{trustmark,vine,robustwide}.

Beyond planar images, watermarking has been extended to meshes~\cite{peng2026pca}, point clouds~\cite{zaman2025deep}, neural radiance fields (NeRF)~\cite{jang2024waterf}, and 3D Gaussian Splatting (3DGS)~\cite{jang20253d}, embedding ownership information in geometry or scene representations. Watermark recovery from rendered views has been explored in 3D-to-2D mesh watermarking~\cite{yoo2022deep} and depth-image-based rendering (DIBR) watermarking~\cite{asikuzzaman2016dibr}, providing related settings for panoramic viewport extraction.

Despite this connection, these approaches do not directly resolve the geometric challenges of panoramic watermarking. Planar methods lack explicit modeling of spherical rotations, while 3D asset methods depend on representations distinct from the dense spherical signal defined on $S^2$. Existing panoramic baselines~\cite{pa1,pa2} rely on geometric remapping or heuristic feature points~\cite{pa3}, which complicates extraction and makes it sensitive to geometric misalignment. Their zero-bit verification also limits multi-bit copyright tracing. Crucially, these panoramic methods do not explicitly account for the intrinsic $SO(3)$ symmetry of panoramic viewing, leaving rotation-robust synchronization and direct viewport extraction unresolved.

\subsection{Learning on Spherical Domains and Equivariance}
To mitigate distortions introduced by planar projections, Spherical CNNs~\cite{eq6} generalize convolutions to the sphere/$SO(3)$ manifold. However, continuous spherical convolutions are computationally expensive. To improve efficiency, approximations based on discrete subgroups~\cite{weiler2019general}, such as the Icosahedral group ($\mathcal{I}_{60}$)~\cite{eq1}, have been explored for tasks including classification and geometric reasoning~\cite{eq2,eq9}. These approaches are grounded in Group Equivariance theory~\cite{cohen2016}, which ensures that transformations in the input space induce predictable transformations in the feature space.
TRIAD~\cite{chen2026triad} embeds watermarks in spherical harmonic coefficients and recovers them using an $SO(3)$-invariant bispectrum. RSWP uses icosahedral facets and a dedicated viewport decoder for direct watermark recovery from rendered views.

\section{Preliminaries}
\subsection{Equivariance}
Equivariance is a fundamental concept in mathematics for characterizing symmetry-preserving mappings. Formally, consider a symmetry group \( \mathcal{G} \), which is defined as a set of transformations that preserve certain structural properties of the spaces \( \mathcal{X} \) and \( \mathcal{Y} \). The group \( \mathcal{G} \) acts on these spaces through transformations \( \mathcal{T}_g \) and \( \mathcal{T}'_g \), respectively, for each \( g \in \mathcal{G} \). A mapping \( m: \mathcal{X} \rightarrow \mathcal{Y} \) is termed equivariant with respect to the group \( \mathcal{G} \) if it satisfies the condition:
\begin{equation}
m(\mathcal{T}_g x) = \mathcal{T}'_g m(x), \quad \forall x \in \mathcal{X}, g \in \mathcal{G}.
\label{equivariance}
\end{equation}
In this context, invariance represents a specific case of equivariance, where the transformation \( \mathcal{T}'_g \) on the output space \( \mathcal{Y} \) is the identity transformation:
\begin{equation}
	m(\mathcal{T}_g x) = m(x), \quad \forall g \in \mathcal{G}.
\end{equation}

\subsection{Icosahedral Rotational Group}
\label{section:ico}
Although functions that are equivariant to $SO(3)$ (the group of 3D rotations) are theoretically desirable for panoramic imagery, directly operating on the continuous group $SO(3)$ is computationally challenging. Therefore, we approximate the equivariance of continuous groups using discrete subgroups. Unlike $SO(2)$, which admits arbitrarily fine cyclic subgroups \(C_{n}\), the largest discrete subgroup of $SO(3)$ that is not contained in a planar rotation group is the icosahedral group $\mathcal{I}_{60}$, composed of 60 group elements. This group can be conceptualized as a regular icosahedron, which has 12 vertices and 20 triangular facets.
\begin{figure}[t] %

	\centering %
	\includegraphics[width=\columnwidth]{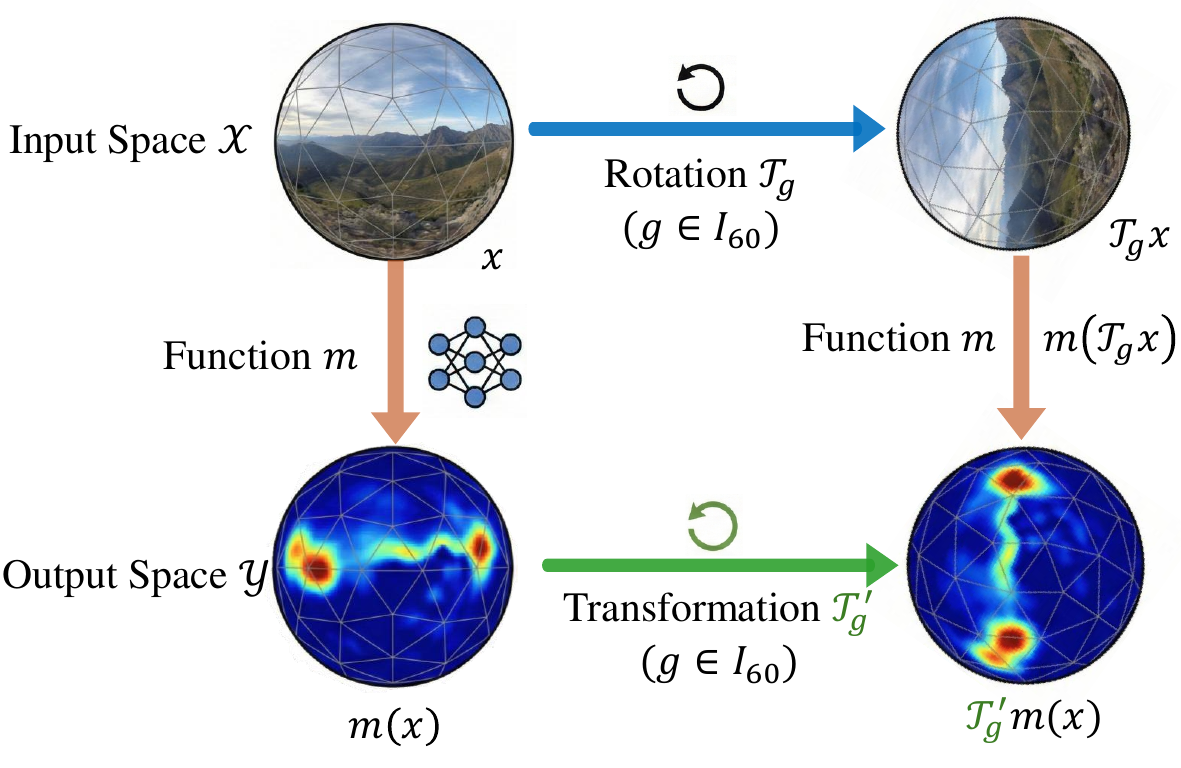} %
	\caption{Illustration of the group equivariance commutative diagram. $m$ is designed to satisfy the equivariance constraint: $m(\mathcal{T}_g x) = \mathcal{T}'_g m(x)$. This guarantees that any input rotation $\mathcal{T}_g$ (where $g \in \mathcal{I}_{60}$) results in a predictable, tractable transformation $\mathcal{T}'_g$ in the feature space, ensuring watermark synchronization without geometric loss.} %
	\label{fig:Icosahedron} %

\end{figure}
\begin{figure*}[ht] %
	\centering %
	\includegraphics[width=\textwidth]{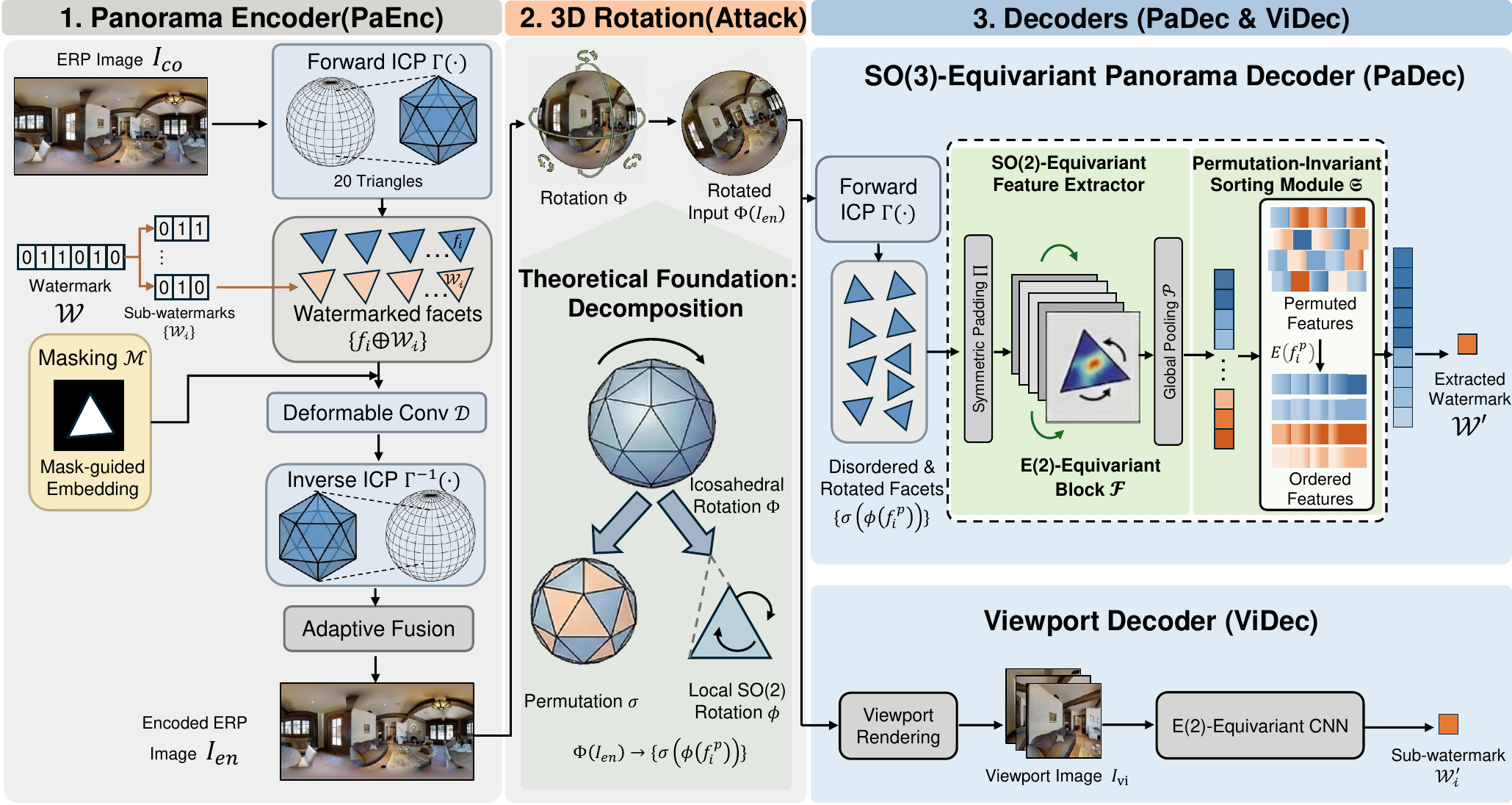} %
  \caption{The overall architecture of the proposed RSWP framework. (a) Decoupled Encoding: The input panoramic signal is projected onto discrete icosahedral facets, allowing the watermark to be embedded in a locally planar, distortion-free domain before being re-projected to the ERP format.
(b) Dual-Path Decoding: The system features two specialized decoding pathways. The $SO(3)$-Equivariant Panorama Decoder (PaDec) recovers watermarks from global rotations by transforming geometric distortions into tractable feature permutations via a sorting module. Simultaneously, the Viewport Decoder (ViDec) exploits the local planar property of the facets to directly extract watermarks from partial 2D renderings without inverse remapping. %
  }
	\label{fig:architecture} %
\end{figure*}

\section{Methodology}

To achieve robustness against spherical distortions, we fundamentally reformulate the watermarking problem through the lens of group equivariance. As defined in Fig.~\ref{fig:Icosahedron}, our objective is to construct a mapping such that a rotation $\mathcal{T}_g$ in the input signal space induces a predictable transformation $\mathcal{T}'_g$ in the feature space. We approximate the continuous sphere using the icosahedral group ($\mathcal{I}_{60}$), the largest discrete subgroup of $SO(3)$. As illustrated in the decomposition step of Fig.~\ref{fig:architecture}, this geometric projection enables us to decouple complex, non-linear spherical rotations into two tractable operations: (1) a combinatorial permutation ($\sigma$) that globally reorders the spatial indices of the triangular facets, mapping a facet from position $i$ to $j$; and (2) a local planar rotation ($\phi$) where the pixel content within each re-indexed facet undergoes a geometric transformation equivalent to a 2D Euclidean rotation ($SO(2)$). Mathematically, this allows us to express the action of an icosahedral rotation on the projected features as a composition: $\Gamma(\Phi(I_{en})) = \{\sigma(\phi(f_i))\}$. Consequently, the challenge of full $SO(3)$ equivariance is converted into a dual problem: maintaining $SO(2)$-equivariance for localized facet features and ensuring permutation invariance for the global facet sequence. The proposed framework leverages this theoretical decomposition to construct a watermarking system inherently robust to 3D rotations by synchronizing these decoupled transformations.

\subsection{Overall Architecture}

Given an ERP image $I_{\mathrm{co}} \in \mathbb{R}^{3 \times H \times W}$, the Icosahedral Projector (ICP) projects it onto the 20 triangular facets of an icosahedron, producing $\{f_i\}_{i=1}^{20}$. A binary watermark $\mathcal{W}$ is partitioned into 20 sub-watermarks $\{\mathcal{W}_{i}\in \mathbb{R}^{\ell_w}\}_{i=1}^{20}$, each assigned to a facet. The Panorama Encoder (PaEnc) embeds each sub-watermark into its corresponding facet and aggregates them to reconstruct the encoded ERP image $I_{\mathrm{en}}$.

During extraction, two decoding pathways operate in parallel. The Panorama Decoder (PaDec) first re-projects the (possibly rotated or attacked) encoded panorama back to icosahedral facets $\{f_{i}^\text{en}\}_{i=1}^{20}$ via ICP, and applies an ${SO}(3)$-equivariant network to extract the sub-watermarks jointly. Simultaneously, the Viewport Decoder (ViDec) processes a rendered viewport image $I_{vi}$ directly from $I_{\mathrm{en}}$, recovering the watermark from the perspective-limited observation. This dual-pathway design ensures both global robustness to 3D rotations and reliable extraction from localized views.

\subsection{Icosahedral Projector (ICP)}
ICP maps an ERP image $I_\text{co}$ of resolution $H\times W$ to twenty triangular facet feature maps $\{f_i\}_{i=1}^{20}$ at a canonical resolution $\mathcal{S}$ via $\Gamma(\cdot)$. The inverse remapping $\Gamma^{-1}(\cdot)$ scatters facet signals back to their ERP coordinates, leaving un-sampled locations as zeros. Algorithm~\ref{alg:icp} specifies the projection and remapping procedures; their detailed description is provided in Section~S7 of the supplementary material.

\begin{algorithm}[t]
\caption{Icosahedral Forward Projection and Inverse Remapping}
\label{alg:icp}
\SetKwInOut{Input}{Input}
\SetKwInOut{Output}{Output}
\SetKwFunction{FProj}{Project}
\SetKwFunction{FInv}{InverseRemap}

\Input{ERP $I_{\text{co}}$ of size $H \times W$; Facet resolution $\mathcal{S}$.}
\Output{Projected facets $\{f_i\}$ / Sparse tensor $\mathbf{F_p}$}

\BlankLine

\textbf{Function} \FProj{$I_{\text{co}}$}:
\Indp
    Initialize list $\{f_i\}$ for $i=1 \dots 20$. \\
    \For{$i \leftarrow 1$ \textbf{to} $20$}{
        Generate local grid $G_i$ of size $\mathcal{S} \times \frac{\sqrt{3}}{2}\mathcal{S}$. \\
        \For{\textbf{each point} $p \in G_i$}{
            $(\theta, \phi) \leftarrow \text{GetSphericalCoords}(p)$ \\
            $(u, v) \leftarrow \text{MapToPixel}(\theta, \phi, H, W)$
            $f_i[p] \leftarrow I_{\text{co}}(u, v)$ \Comment{Bilinear Interpolation}
        }
    }
    \Return $\{f_i\}_{i=1}^{20}$
\Indm

\BlankLine

\textbf{Function} \FInv{$\{f_i\}_{i=1}^{20}, H, W$}:
\Indp
    Initialize $\mathbf{F_p}$ of size $C \times H \times W$ with zeros. \\
    \For{$i \leftarrow 1$ \textbf{to} $20$}{
        \For{\textbf{each point} $p \in G_i$}{
            $(\theta, \phi) \leftarrow \text{GetSphericalCoords}(p)$ \\
            $(u, v) \leftarrow \text{MapToPixel}(\theta, \phi, H, W)$ \\
            $\mathbf{F_p}[u, v] \leftarrow f_i[p]$ \Comment{Sparse Scatter}
        }
    }
    \Return $\mathbf{F_p}$
\Indm
\end{algorithm}

\subsection{Panorama Encoder (PaEnc)}
Panorama Encoder is designed to embed the partitioned sub-watermarks $\{\mathcal{W}_{i}\in \mathbb{R}^{\ell_w}\}_{i=1}^{20}$ into the projected triangular facets $\{f_i\}_{i=1}^{20}$. By embedding information into the icosahedral domain, the watermark bits are naturally and redundantly distributed across spatially separated regions. This topological distribution ensures that any rendered viewport image contains recoverable signal fragments, regardless of the viewing angle. The encoding process consists of two stages: \textit{Mask-Guided Facet Embedding} and \textit{Spatially Adaptive Fusion}.

\vspace{0.5em}\noindent\textbf{Mask-Guided Facet Embedding.} This initial stage embeds the watermark into individual facets while alleviating the domain mismatch between triangular signals and rectangular CNN receptive fields. For each facet $f_i \in \mathbb{R}^{C \times \mathcal{S} \times \text{H}_f}$, we expand the binary sub-watermark $\mathcal{W}_i$ to the facet resolution and concatenate it along the channel dimension. This results in a composite tensor of shape $(C + \ell_w , \mathcal{S} , \text{H}_f)$. To strictly confine the embedding to the valid geometric region, we apply a triangular mask $\mathcal{M}$ after concatenation. This ensures that the zero-padded exterior regions do not contribute to the feature embedding, thereby precluding boundary artifacts. The masked tensor is then processed by a stack of Deformable Convolution layers ($\mathcal{D}$) to accommodate the geometric distortions inherent in the spherical-to-planar projection by adaptively adjusting sampling locations. The embedding process for the $i$-th facet is formulated as:\begin{equation}\mathbf{F}^i = \mathcal{D} \left( \mathcal{M} \odot \left( f_i \oplus \mathcal{W}_i \right) \right),\end{equation}where $\oplus$ denotes channel-wise concatenation and $\odot$ represents element-wise multiplication. The output $\mathbf{F}^i$ serves as the deep feature representation of the watermarked facet.

\vspace{0.5em}\noindent\textbf{Spatially Adaptive Fusion.} Once the facet embeddings $\{\mathbf{F}^i\}_{i=1}^{20}$ are generated, they are mapped back to the global panoramic view. We employ the inverse remapping of ICP $\Gamma^{-1}(\cdot)$ to re-project the deep features from the disjoint facets onto the ERP domain:\begin{equation}\mathbf{F_p} = \Gamma^{-1} \left( { \mathbf{F}^i  }_{i=1}^{20} \right).\end{equation}This operation results in a sparse feature map $\mathbf{F_p} \in \mathbb{R}^{C \times H \times W}$, where valid features exist only at the projected coordinate locations. To fuse these features into the original image imperceptibly, we introduce a learnable Spatially Adaptive Weighting matrix $\mathbf{W}_{\text{adapt}} \in \mathbb{R}^{C \times H \times W}$. This matrix modulates the embedding strength pixel-wise, learning to assign lower weights to smooth regions (where artifacts are visible) and higher weights to textured regions. The weighted features are then compressed to 3 channels via a $1\times1$ convolution $\mathbf{Conv}_{C\to3}$ to generate the final watermark residual $\mathbf{F_w}$:
\begin{equation}
\mathbf{F_w} = \mathbf{Conv}_{C\to3} \left( \mathbf{F_p} \odot \mathbf{W}_{\text{adapt}} \right).
\end{equation}
Finally, this residual is added to the original cover image $I_{\text{co}}$ to produce the encoded panorama:
\begin{equation}
I_{\text{en}} = I_{\text{co}} + \mathbf{F_w}.
\end{equation}

\vspace{0.5em}
\textbf{Loss Functions.}
To balance embedding strength and maintain perceptual fidelity, we define an embedding loss $\mathcal{L}_{E}$ consisting of the Mean Squared Error (MSE) computed in both the global ERP domain and the local individual facets before and after embedding:
\begin{equation}
	\mathcal{L}_{E} = \text{MSE}(I_{\text{co}}, I_{\text{en}}) + \sum_{i=1}^{20} \text{MSE}(f_i, f_i^{\text{en}}),
\end{equation}
where $f_i^{\text{en}}$ denotes the $i$-th encoded facet.

\subsection{SO(3)-Equivariant Panorama Decoder (PaDec)}

\paragraph{Projection to Icosahedral Facets}
Given an encoded panoramic image $I_{\mathrm{en}}$, the icosahedral projector maps it to a set of 20 triangular facets:
\begin{equation}
\Gamma(I_{\mathrm{en}}) = \{ f_i^{\mathrm{en}} \}_{i=1}^{20}.
\end{equation}
For any icosahedral rotation $\Phi \in \mathcal{I}_{60}$, the action of $\Phi$ on the global spherical signal is mathematically equivalent to a permutation
$\sigma$ of facet indices, together with an in-plane rotation $\phi$ applied to each facet.
Thus, the rotated facets satisfy:
\begin{equation}
\label{eq:icosahedron}
\Gamma(\Phi(I_{\mathrm{en}}))
= \{ \tilde{f}_j^{\mathrm{en}} \}_{j=1}^{20}
= \{\, \sigma(\phi(f_i^{en})) \,\}_{i=1}^{20},
\end{equation}
where $\tilde{f}_j^{\mathrm{en}}$ denotes the facet occupying the $j$-th position post-rotation.
This representation explicitly separates the in-plane transformation $\phi$ from the inter-facet permutation $\sigma$.

\paragraph{Symmetric Padding of Triangle Facet} We apply symmetric padding $\Pi$ to each triangular facet $f_i$ to obtain the canonical representation $f_i^p$ used by the equivariant feature extractor. The geometric explanation and illustration are provided in Section~S5 of the supplementary material; the construction is specified in Algorithm~\ref{alg:rotation_padding}.

\begin{algorithm}[t]
\caption{Symmetric Padding $\Pi$}
\label{alg:rotation_padding}
\SetKwInput{Input}{Input}
\SetKwInput{Output}{Output}

\Input{Triangular facet $f_i$ with side length $S$}
\Output{Symmetrically padded facet $f_i^p \in \mathbb{R}^{2r \times 2r}$}

$r \gets S / \sqrt{3}$ \Comment{Derive canonical radius}\\
$L \gets 2r$ \Comment{Target square side length}\\
$H_f \gets \frac{\sqrt{3}}{2}S$ \Comment{Triangle height}\\

$p_{lr} \gets (L - S) / 2$ \Comment{Symmetric horizontal padding}\\

\tcc{Align centroid to square center}
\uIf{$\operatorname{Orientation}(f_i) = \text{UPRIGHT} \ (\triangle)$}{
    $p_{\mathrm{top}} \gets 0$ \\
    $p_{\mathrm{bottom}} \gets L - H_f$
}
\Else(\tcc*[f]{$\operatorname{Orientation}(f_i) = \text{INVERTED} \ (\nabla)$}){
    $p_{\mathrm{top}} \gets L - H_f$ \\
    $p_{\mathrm{bottom}} \gets 0$
}

$f_i^p \gets \operatorname{Pad}(f_i,\, \text{padding}=(p_{\mathrm{top}},\, p_{\mathrm{bottom}},\, p_{\mathrm{lr}},\, p_{\mathrm{lr}}))$ \\
\end{algorithm}

\paragraph{SO(2)-Equivariant Feature Extractor and Pooling}
Each facet $f_i^{p}$ is independently processed by an E(2)-equivariant ResNet block $\mathcal{F}$.
A global mean pooling operator $\mathcal{P}$ is then applied to eliminate any residual in-plane rotational dependence.

\begin{proposition}[Equivariance to Icosahedral Rotation of the ResNet Block]
Let $\mathcal{F}$ be an E(2)-equivariant ResNet block and $\mathcal{P}$ a global mean pooling operator.
For any icosahedral rotation $\Phi$ acting as $\{\sigma(\phi(f_i^{p}))\}$, it holds that:
\begin{equation}
\mathcal{P}\mathcal{F}(\{\sigma(\phi(f_i^{p}))\}_{i=1}^{20})
=
\sigma(\mathcal{P}\mathcal{F}(\{f_i^{p}\}_{i=1}^{20})),
\end{equation}
which is consistent with the Equation (1).
\end{proposition}

\begin{proof}
As stated in Equation (\ref{eq:icosahedron}), any
icosahedral rotation $\Phi$ can be decomposed into $\phi$ and $\sigma$.
Let $\mathcal{F}$ be composed of group convolutions with kernels constrained to be ${SO}(2)$-steerable.
By the definition of group convolution on $G={SO}(2)$, convolving a rotated input $\phi(f_i^{p})$
with a steerable kernel is equivalent to rotating the convolved output:
\begin{equation}
(\kappa * \phi(f_i^{p}))(x)
=
(\phi\kappa * f_i^{p})(x)
=
\phi\big((\kappa * f_i^{p})(x)\big).
\end{equation}
Since residual connections and nonlinearities in E(2)-equivariant networks~\cite{weiler2019general} preserve equivariance,
the property holds layer-wise and thus for the entire block $\mathcal{F}$.
Therefore, for any in-plane rotation $\phi$ we have:
\begin{equation}
\mathcal{F}(\phi(f_i^{p})) = \phi(\mathcal{F}(f_i^{p})).
\end{equation}
Applying group pooling $\mathcal{P}$ further removes the rotational dependence:
\begin{equation}
\mathcal{P}(\phi(x)) = \mathcal{P}(x).
\end{equation}
Finally, as $\sigma$ merely permutes facet indices and $\mathcal{F},\mathcal{P}$ operates independently on each facet,
the equality follows.
\end{proof}

\paragraph{Permutation-Invariant Sorting Module}
To eliminate dependence on the residual permutation $\sigma$, we introduce a permutation-invariant sorting operator $\mathfrak{S}$. Let
\begin{equation}
\begin{aligned}
     \mathbb{E}_{\text{C},\text{H}_f, \mathcal{S}}(f_i^{p}) &= \mathbb{E}_{\text{C},\text{H}_f, \mathcal{S}}(\phi(f_i^{p})) , \quad \forall i \in [1..20]\\
     &=\frac{1}{\text{H}_{f} \times \mathcal{S}\times \text{C}} \sum_{t=1}^{\text{H}_f}
     \sum_{j=1}^\mathcal{S}\sum_{k=1}^\text{C} (f_i^{p})_{t,j,k},
\end{aligned}
\end{equation}
be the average of all pixels across channels and spatial dimensions. The operator $\mathfrak{S}$ sorts $\{f_i^{p}\}$ by $\mathbb{E}(f_i^{p})$,
\begin{equation}
     \mathfrak{S}\left(\ \{f_i^{p}\}_{i=1}^{20}\right)
     = \xi\!\left(\ {\mathbb{E}_{\text{C},\text{H}_f, \mathcal{S}}}\left(\ \{f_i^{p}\}_{i=1}^{20}\right)\right),
\end{equation}
where $\xi(\cdot)$ denotes the sorting operation.

\begin{proposition}[Permutation Invariance of $\mathfrak{S}$]
For any permutation $\sigma$ and facet set $\{f_i^{p}\}$,
\begin{equation}
\mathfrak{S}(\{\sigma(f_i^{p})\}_{i=1}^{20}) = \mathfrak{S}(\{f_i^{p}\}_{i=1}^{20}).
\end{equation}
\end{proposition}

\begin{proof}
For any $\sigma$, the set of averages remain unchanged:
\begin{equation}
\{\mathbb{E}(\sigma(f_i^{p}))\} = \{\mathbb{E}(f_i^{p})\}.
\end{equation}
Sorting identical scalar values yields identical ordered outputs:
\begin{equation}
\mathfrak{S}\left(\ \{\sigma(\phi(f_i^{p}))\}_{i=1}^{20}\right)
= \rho\left(\ \{\phi(f_i^{p})\}_{i=1}^{20}\right),
\end{equation}
\begin{equation}
\mathfrak{S}\left(\ \{f_i^{p}\}_{i=1}^{20}\right)
= \rho\left(\ \{f_i^{p}\}_{i=1}^{20}\right),
\end{equation}
where $\rho$ maps $f_i^{p}$ and $\phi(f_i^{p})$ to the same order.
\end{proof}

\paragraph{Theoretical Guarantee of Icosahedral Invariance}

\begin{theorem}[Icosahedral Rotation Invariance of the Extractor]
The composed network $\mathcal{N} = \mathcal{P}\mathcal{F}\mathfrak{S}\Gamma(\cdot)$ is invariant under any icosahedral rotation $\Phi\in\mathcal{I}_{60}$, i.e.
\begin{equation}
\mathcal{N}(\Phi(I_\text{en})) = \mathcal{N}(I_\text{en}).
\end{equation}
\end{theorem}

\begin{proof}
Applying $\Phi$ yields $\{\sigma(\phi(f_i))\}$. By Proposition 1,
$\mathcal{P}\mathcal{F}(\{\sigma(\phi(f_i))\}) = \sigma(\mathcal{P}\mathcal{F}(\{f_i\}))$. By Proposition 2, $\mathfrak{S}(\sigma(\cdot)) = \mathfrak{S}(\cdot)$. Therefore, the composed output is invariant to $\Phi$:
\begin{equation}
\begin{aligned}
\mathcal{P}\mathcal{F}\mathfrak{S}\Gamma(\Phi(I_\text{en})) &= \mathcal{P}\mathcal{F}\mathfrak{S}(\{{\tilde{f}}^{\text{en},s}_i\}_{i=1}^{20}) \\
&=  \mathcal{P}\mathcal{F}\mathfrak{S}(\{\sigma(\phi({f}^{\text{en},s}_i))\}_{i=1}^{20})\\
&= \mathcal{P}\mathcal{F}\rho(\{\phi({f}^{\text{en},s}_i)\}_{i=1}^{20})\\
&= \mathcal{P}\mathcal{F}\rho(\{{f}^{\text{en},s}_i\}_{i=1}^{20})\\
&= \mathcal{P}\mathcal{F}\mathfrak{S}\Gamma(I_\text{en}).
\end{aligned}
\end{equation}
\end{proof}

As the objective of PaDec is to recover a watermark $\mathcal{W'}$ that matches the embedded message $\mathcal{W}$, the loss ${\mathcal{L}}_{Pa}$ is defined as the MSE between $\mathcal{W}$ and $\mathcal{W'}$:
\begin{equation}
	\mathcal{L}_{Pa} = \text{MSE}(\mathcal{W}, \mathcal{W'}).
\end{equation}

\begin{figure}[t] %

	\centering %
	\includegraphics[width=\columnwidth]{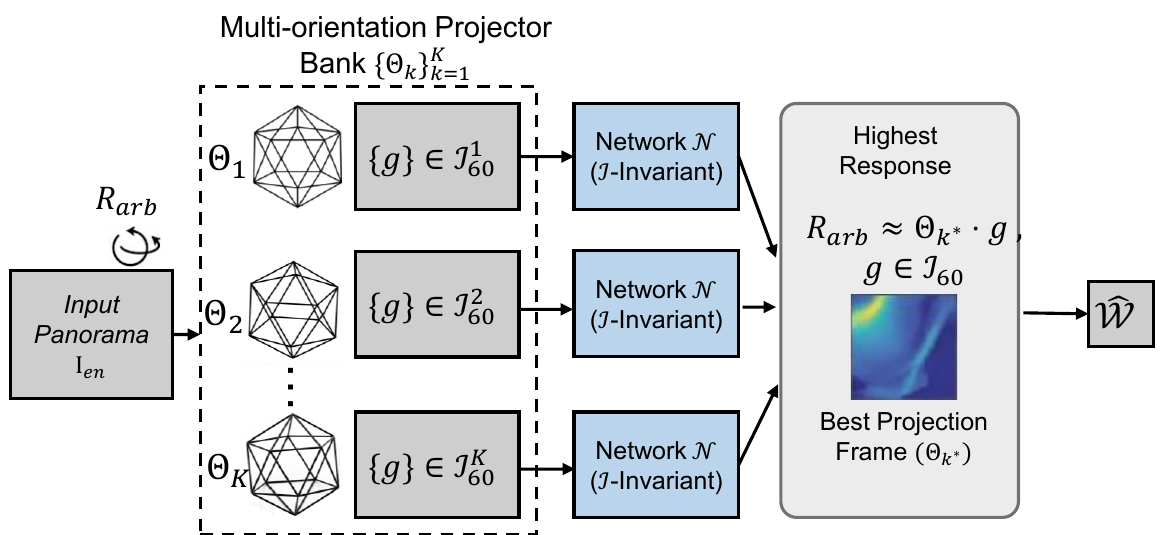} %
	\caption{The Multi-Orientation Projector Bank. To bridge the gap between discrete $\mathcal{I}_{60}$-invariance and continuous world rotations, the bank aligns arbitrary inputs to the nearest canonical frame among $K$ presets.}
	\label{fig:fullrotation} %

\end{figure}

\subsection{Continuous SO(3) Robustness via Projector Bank}
While Theorem 1 establishes strict invariance to the discrete icosahedral group $\mathcal{I}_{60}$, real-world distortions involve continuous rotations $R \in {SO}(3)$. To bridge the gap between discrete group invariance and continuous robustness, we introduce a \textit{Multi-Orientation Projector Bank} $\mathcal{B}=\{\Theta_k\}_{k=1}^K$ that densely covers the quotient space ${SO}(3) / \mathcal{I}_{60}$.
As shown in Fig.~\ref{fig:fullrotation}, we pre-define a set of $K$ canonical orientations $\{\Theta_k\}_{k=1}^K$ for the Icosahedral Projector.
For any arbitrary rotation $R_{\text{arb}}$, there exists at least one preset orientation $\Theta_{k^*}$ that minimizes the residual misalignment (i.e., $R_{\text{arb}} \approx \Theta_{k^*} \cdot g$ for some $g \in \mathcal{I}_{60}$).
Crucially, these orientations are implemented as precomputed coordinate lookup tables. Switching between orientations $\Theta_k$ only requires changing the sampling grid indices of ICP, incurring negligible geometric transformation overhead. Although each candidate orientation is processed through the extractor $\mathcal{N}$, the evaluation is fully parallelizable and therefore suitable for offline verification.
During extraction, we employ a confidence-based selection metric. Let $D_l(\cdot)$ denote the output logits of the decoder. We identify the optimal orientation $\hat{\Theta}$:
\begin{equation}
\hat{\Theta} = \operatorname*{arg\,max}_{\Theta_k \in \mathcal{B}} \frac{1}{L} \sum |D_l(\Gamma(I_{en}, \Theta_k))|
\end{equation}
where $|\cdot|$ denotes the absolute magnitude. A higher response magnitude indicates superior alignment with the learned watermarking features, serving as a reliable proxy for extraction.

\textbf{Construction of the Orientation Bank.}
To guarantee a worst-case angular misalignment bounded by $\epsilon = 5^\circ$, we construct the orientation bank $\mathcal{B} = \{\Theta_k\}$ by sampling the quotient space ${SO}(3) / \mathcal{I}_{60}$.
The icosahedral group $\mathcal{I}_{60}$ partitions ${SO}(3)$ into 60 congruent fundamental domains (Voronoi cells). Let $\Omega$ denote one such fundamental domain. The maximum geodesic distance from the identity element to the boundary of $\Omega$ is approximately $22.2^\circ$.
Using a greedy furthest-point sampling strategy, we generate an optimal $\epsilon$-net that guarantees  for any rotation $R \in$ $\Omega$, there exists a $\Theta_k \in \mathcal{B}$ satisfying $d(R, \Theta_k) \le \epsilon$, where $d(\cdot, \cdot)$ denotes the geodesic distance on ${SO}(3)$. This mathematically ensures that any continuous rotation is quantized into robust zone.

\subsection{Viewport Decoder (ViDec)}
For rendered viewport images, we train an additional decoder to directly extract watermarks while mitigating rendering-induced distortions. Since a viewport image is naturally a planar 2D image, we employ the same E(2)-equivariant CNN architecture as PaDec, enabling ViDec to maintain equivariance to ${SO}(2)$.
During training, viewport images that cover watermarks $\mathcal{W}_i, \forall i \in [1..20]$ from various icosahedral facets are provided as inputs. The decoder is trained to recover the corresponding watermark $\mathcal{W}'_{i}, \forall i \in [1..20]$ within the associated facet. The loss is defined as:
\begin{equation}
	\mathcal{L}_{Vi} = \text{MSE}(\mathcal{W}_i, \mathcal{W}'_{i}), \forall i \in [1..20].
\end{equation}

\subsection{Training Strategy}
\label{sec:training}
The overall loss function of  PaEnc, PaDec and ViDec is:
\begin{equation}
	\mathcal{L} = {\lambda}_{E}\mathcal{L}_{E}+{\lambda}_{Pa}\mathcal{L}_{Pa}+{\lambda}_{Vi}\mathcal{L}_{Vi},
\end{equation}
where ${\lambda}_{E}, {\lambda}_{Pa}~\text{and}~\lambda_{Vi}$ are the weighting factors. Specifically, the training is separated into two branches: PaEnc-PaDec and PaEnc-ViDec. For the first 5 epochs, only PaEnc-PaDec is trained to allow PaEnc-PaDec to adapt to the structure of ERP, and for the next 5 epochs, only PaEnc-ViDec is trained with PaDec being fixed in order to focus on the viewport messages. Later the two branches are trained alternately for 50 epochs.
\section{Experiments}
\textbf{Dataset.} We use panoContext \cite{zhang2014panocontext} and SUN360\cite{sun} as experimental datasets. 8,000 images are randomly selected for training set and 2,000 for testing. The size of each panoramic image is $ 640 \times 320$, and that of each viewport image is  $128 \times 128$ with a latitude span of $2\pi/5$ and a longitude span of $\pi/2$.

\textbf{Experimental Settings.} All experiments are implemented on an NVIDIA A100 GPU. The resolution of ICP is $\mathcal{S} =128$ and the size of pre-defined Orientation bank is $K=500$ with an average inference time of $0.15s$. The watermark message is a randomly generated 320-bit sequence. The E(2)-equivariant  network $\mathcal{F}$ is implemented using ESCNN~\cite{weiler2019general}. We utilize Adam as the optimizer with a learning rate of $1\times10^{-4}$ . During training, we initially set ${\lambda}_{E}=0.6$, ${\lambda}_{Pa}=2$ and ${\lambda}_{Vi}=2 $. After 20 epochs,  ${\lambda}_{E}$ is increased by 0.2 per epoch, ensuring PaEnc and ViDec achieves sufficient extraction accuracy before improving visual quality. The model is trained for 60 epochs in total.

\textbf{Baseline Methods.} The comparative watermarking methods are twofold. We include the recent state-of-the-art watermarking approaches designed for conventional 2D images, including MBRS \cite{mbrs}, Robust-Wide \cite{robustwide}, TrustMark \cite{trustmark}, and VINE \cite{vine}. The second category includes JND \cite{pa2}, a watermarking method specifically for panoramic images.

\textbf{Evaluation Metrics.} We use PSNR, SSIM, and LPIPS~\cite{zhang2018unreasonable} to evaluate the visual fidelity. For robustness, we report the average Panoramic Bit Accuracy (PBA) and Viewport Bit Accuracy (VBA), which are the ratio of correctly extracted watermark bits to the embedded bits. Furthermore, it is essential to strictly control the false positive rate (FPR), as incorrectly labeling non-watermarked images as watermarked can be detrimental. Thus, we measure the true positive rate (TPR) under low false positive rate (FPR) levels for panoramas (PTPR@ 0.1\% FPR) and viewports(VTPR@ 0.1\% FPR).

\begin{figure*}[ht]
	\centering
   \includegraphics[width=\textwidth,trim=0 624bp 0 0,clip]{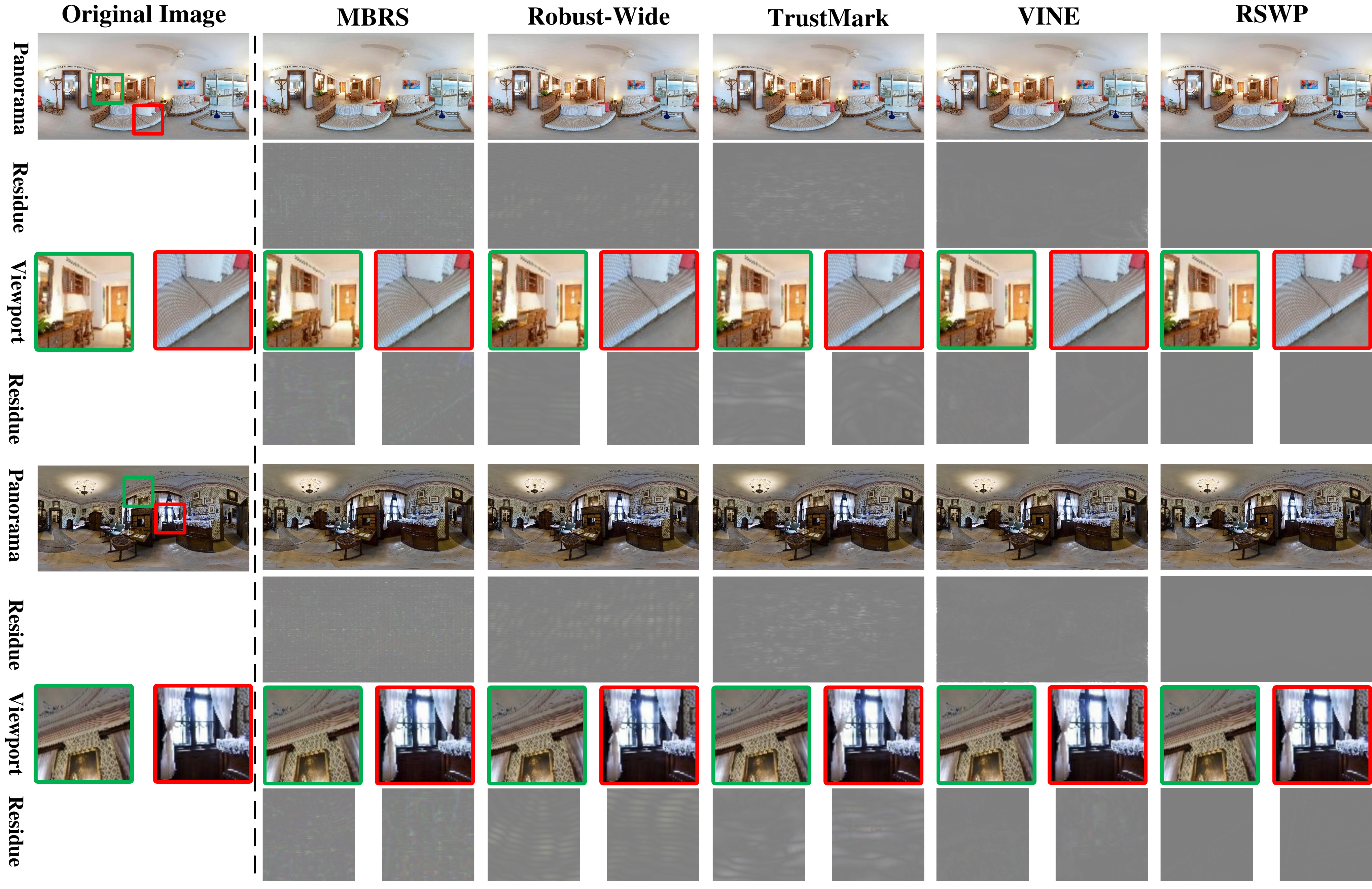}
	\caption{Comparison of watermarked panorama  and viewport images between RSWP and baseline methods.}
	\label{fig:compare_vi}
\end{figure*}

\subsection{Visual Fidelity Analysis}
Table~\ref{tab:combined} presents the quantitative comparison of visual quality between RSWP and baseline methods. RSWP consistently outperforms all competing approaches across PSNR, SSIM, and LPIPS metrics.

As shown in Fig.~\ref{fig:compare_vi}, the superior fidelity of our method stems from the resolution-decoupling strategy inherent in Algorithm~\ref{alg:icp}. Traditional watermarking schemes typically embed signals densely across the entire spatial domain, which requires widespread pixel perturbations that degrade visual quality across the whole ERP. Furthermore, a higher watermark capacity leads to more pronounced degradation.

In contrast, RSWP effectively decouples the embedding resolution $\mathcal{S}$ from the input panoramic resolution $H \times W$. The watermark is embedded within the compact, lower-resolution icosahedral facet domain. Crucially, during the reconstruction phase, the inverse remapping $\Gamma^{-1}(\cdot)$ projects these watermarked features back onto the high-resolution ERP domain. This scattering mechanism ensures that the watermark signal is injected only at sparse, discrete spatial coordinates rather than perturbing the continuous pixel grid especially those noise-sensitive regions. Consequently, the vast majority of the original high-frequency details remain structurally intact, fundamentally minimizing distortion while maintaining a robust global signal structure suitable for blind extraction.

\begin{table}[h]
    \centering
    \caption{Visual fidelity comparison for different methods.}
    \begin{adjustbox}{max width=\columnwidth}
    \begin{tabular}{l|c|ccc|ccc}
        \toprule
        \multirow{2}{*}{\textbf{Method}}
        & \multirow{2}{*}{\textbf{Capacity}}
        & \multicolumn{3}{c|}{\textbf{Panorama}}
        & \multicolumn{3}{c}{\textbf{Viewport}} \\
        \cmidrule{3-8}
        & & \textbf{PSNR (dB)$\uparrow$} & \textbf{SSIM$\uparrow$} & \textbf{LPIPS$\downarrow$}
          & \textbf{PSNR (dB)$\uparrow$} & \textbf{SSIM$\uparrow$} & \textbf{LPIPS$\downarrow$} \\
        \midrule
        MBRS\cite{mbrs} & 256 & 36.70 & 0.9194 & 0.0208 & 37.00 & 0.9665 & 0.0095 \\
        TrustMark\cite{trustmark} & 100 & 41.56 & 0.9914 & 0.0019 & 39.68 & 0.9920 & 0.0016 \\
        Robust-Wide\cite{robustwide} & 64 & 43.69 & 0.9844 & 0.0032 & 43.63 & 0.9800 & 0.0046 \\
        VINE\cite{vine} & 100 & 39.46 & 0.9876 & 0.0013 & 43.26 & 0.9931 &0.0008\\
        JND\cite{pa2} &1 & 43.35 & 0.9651 & 0.0048 & 42.38 & 0.9553 & 0.0043 \\
       \rowcolor[gray]{0.9}
        RSWP & \textbf{320} & \textbf{45.86} & \textbf{0.9938} & \textbf{0.0015}
             & \textbf{46.91} & \textbf{0.9942} & \textbf{0.0022} \\
        \bottomrule
    \end{tabular}
    \end{adjustbox}
    \vspace{-0.2cm}
    \label{tab:combined}
\end{table}

\begin{table*}[t]
\centering
\small
\caption{Comparison under icosahedral rotation and arbitrary rotation attacks. Entries are PBA(\%) / TPR@0.1\%FPR. Rotation angles are in (Pitch, Yaw, Roll) format.}
\label{tab:comparison1}
\begin{adjustbox}{max width=\textwidth}
\begin{tabular}{lcccccccccc}
\toprule
\multirow{2}{*}{\textbf{Method}}
 & \multicolumn{6}{c}{\textbf{Rotation of Icosahedral group $\mathcal{I}_{60}$ (PBA$\uparrow$/PTPR@0.1\%FPR$\uparrow$)}}
 & \multicolumn{3}{c}{\textbf{Arbitrary Rotation (PBA$\uparrow$/PTPR@0.1\%FPR$\uparrow$)}} \\
\cmidrule(lr){2-7}\cmidrule(lr){8-10}
 & (0$^{\circ}$, 0$^{\circ}$, 0$^{\circ}$) & (0$^{\circ}$, 144$^{\circ}$, 0$^{\circ}$) & (0$^{\circ}$, 72$^{\circ}$, 0$^{\circ}$) & (180$^{\circ}$, 0$^{\circ}$, 0$^{\circ}$) & (180$^{\circ}$, 72$^{\circ}$, 0$^{\circ}$) & (0$^{\circ}$, 0$^{\circ}$, 180$^{\circ}$)
 & (90$^{\circ}$, 60$^{\circ}$, 30$^{\circ}$)
 & (60$^{\circ}$, 0$^{\circ}$, 60$^{\circ}$)
 & (20$^{\circ}$, 40$^{\circ}$, 15$^{\circ}$) \\
\midrule

MBRS \cite{mbrs}
 & 100.00 / 1.000 & 49.23 / 0.000 & 48.97 / 0.000 & 46.53 / 0.000 & 49.36 / 0.000 & 49.31 / 0.000
 & 48.61 / 0.000 & 47.25 / 0.000 & 49.04/ 0.000 \\

TrustMark \cite{trustmark}
 & 100.00 / 1.000 & 48.11 / 0.000 & 49.97 / 0.001 & 45.66 / 0.000 & 48.21 / 0.000 & 49.52 / 0.001
 & 49.23 / 0.001 & 46.53 / 0.000 & 49.36 / 0.001 \\

Robust-Wide \cite{robustwide}
 & 100.00 / 1.000 & 52.41 / 0.003 & 53.38 / 0.004 & 51.01 / 0.001 & 50.25 / 0.001 & 50.43 / 0.001
 & 50.33 / 0.001 & 51.24 / 0.001 & 50.03 / 0.001 \\

VINE \cite{vine}
 & 100.00 / 1.000 & 49.34 / 0.001 & 51.24 / 0.002 & 49.28 / 0.001 & 50.01 / 0.001 & 50.23 / 0.001
 & 48.26 / 0.000 & 47.84 / 0.000 & 51.39/ 0.001 \\

JND \cite{pa2}
 & - / 1.000 & - / 0.002 & - / 0.001 & - / 0.002 & - / 0.000 & - / 0.001
 & - / 0.002 & - / 0.000 & - / 0.000 \\

\rowcolor[gray]{0.9}
\textbf{RSWP (Ours)}
 & \textbf{100.00 / 1.000} & \textbf{99.65 / 1.000} & \textbf{99.45 / 1.000} & \textbf{99.87 / 1.000}
 & \textbf{99.09 / 0.997} & \textbf{98.65 / 1.000}
 & \textbf{88.74 / 1.000} & \textbf{89.65 / 1.000} & \textbf{83.37 / 0.997} \\
\bottomrule
\end{tabular}
\end{adjustbox}
\end{table*}

\begin{table*}[t]
    \centering
    \small
    \caption{PBA (\%) and VBA (\%) for comparative watermarking methods under different attacks. }
    \label{tab:comparison_attacks}
    \begin{adjustbox}{max width=\textwidth}
    \begin{tabular}{lccccc|ccccc}
        \toprule
        \multirow{2}{*}{\textbf{Attack}}
        & \multicolumn{5}{c|}{\textbf{PBA(\%)$\uparrow$}}
        & \multicolumn{5}{c}{\textbf{VBA(\%)$\uparrow$}} \\
        \cmidrule(lr){2-6}\cmidrule(lr){7-11}
         & MBRS\cite{mbrs}
         & TrustMark\cite{trustmark}
         & Robust-Wide\cite{robustwide}
         & VINE\cite{vine}
         & \cellcolor{gray!20}\textbf{RSWP}
         & MBRS\cite{mbrs}
         & TrustMark\cite{trustmark}
         & Robust-Wide\cite{robustwide}
         & VINE\cite{vine}
         & \cellcolor{gray!20}\textbf{RSWP} \\
        \midrule

        Brightness -50\%
        & 99.59 & 96.55 & 99.89 & 97.05 & \cellcolor{gray!20}\textbf{97.73}
        & 50.82 & 49.94 & 51.63 & 49.12 & \cellcolor{gray!20}\textbf{97.45} \\

        Brightness +50\%
        & 99.27 & 96.23 & 93.19 & 96.89 & \cellcolor{gray!20}\textbf{97.79}
        & 50.47 & 49.71 & 50.92 & 49.36 & \cellcolor{gray!20}\textbf{97.48} \\

        Contrast -50\%
        & 99.36 & 98.23 & 99.51 & 99.52 & \cellcolor{gray!20}\textbf{97.22}
        & 49.93 & 49.15 & 50.28 & 49.01 & \cellcolor{gray!20}\textbf{97.03} \\

        Contrast +50\%
        & 98.24 & 98.16 & 99.02 & 99.46 & \cellcolor{gray!20}\textbf{97.36}
        & 49.58 & 48.94 & 49.84 & 48.73 & \cellcolor{gray!20}\textbf{97.23} \\

        Gaussian Blur $k=1$
        & 100.00 & 99.88 & 100.00 & 99.55 & \cellcolor{gray!20}\textbf{99.98}
        & 50.71 & 50.23 & 51.42 & 49.63 & \cellcolor{gray!20}\textbf{99.49} \\

        Gaussian Blur $k=3$
        & 99.91 & 98.76 & 99.93 & 96.51 & \cellcolor{gray!20}\textbf{99.79}
        & 49.84 & 49.31 & 49.92 & 48.87 & \cellcolor{gray!20}\textbf{98.97} \\

        Gaussian Blur $k=5$
        & 99.68 & 95.41 & 77.19 & 91.23 & \cellcolor{gray!20}\textbf{88.58}
        & 48.73 & 48.19 & 48.64 & 47.92 & \cellcolor{gray!20}\textbf{90.66} \\

        Salt \& Pepper (0.01)
        & 99.98 & 98.75 & 100.00 & 99.89 & \cellcolor{gray!20}\textbf{98.73}
        & 49.63 & 49.22 & 49.87 & 48.81 & \cellcolor{gray!20}\textbf{98.06} \\

        Dropout (0.1)
        & 96.96 & 94.65 & 96.78 & 98.92 & \cellcolor{gray!20}\textbf{96.78}
        & 49.44 & 48.87 & 49.73 & 48.62 & \cellcolor{gray!20}\textbf{96.37} \\

        Cropout (p=10\%)
        & 99.52 & 99.73 & 99.79 & 99.77 & \cellcolor{gray!20}\textbf{99.85}
        & 49.82 & 49.34 & 50.21 & 49.09 & \cellcolor{gray!20}\textbf{99.96} \\

        Cropout (p=20\%)
        & 97.28 & 98.56 & 99.56 & 96.53 & \cellcolor{gray!20}\textbf{99.06}
        & 49.14 & 48.72 & 49.31 & 48.41 & \cellcolor{gray!20}\textbf{99.73} \\

        JPEG (50)
        & 99.01 & 97.63 & 99.85 & 98.98 & \cellcolor{gray!20}\textbf{98.73}
        & 49.07 & 48.69 & 49.41 & 48.54 & \cellcolor{gray!20}\textbf{97.07} \\

        JPEG (80)
        & 99.89 & 98.52 & 99.97 & 99.77 & \cellcolor{gray!20}\textbf{98.73}
        & 50.31 & 49.84 & 50.62 & 49.44 & \cellcolor{gray!20}\textbf{98.55} \\

        Gaussian Noise $\sigma=0.04$
        & 99.79 & 98.97 & 99.89 & 99.06 & \cellcolor{gray!20}\textbf{99.52}
        & 48.91 & 48.34 & 48.87 & 48.02 & \cellcolor{gray!20}\textbf{99.65} \\

        \midrule
        Rotation(180,0,0) + JPEG(80)
        & 48.50 & 48.50 & 49.73 & 53.51 & \cellcolor{gray!20}\textbf{90.67}
        & 46.54  & 47.21 & 49.62 & 48.65 & \cellcolor{gray!20}\textbf{95.63} \\

        (0,144,180) + Cropout(20\%)
        & 44.97 & 44.97 & 53.38 & 51.24 & \cellcolor{gray!20}\textbf{95.45}
        & 50.23 & 48.32 & 41.63 & 48.99 & \cellcolor{gray!20}\textbf{97.32} \\

        Rotation(120,0,0) + Gaussian Noise(0.03)
        & 48.03 & 48.03 & 50.25 & 50.01 & \cellcolor{gray!20}\textbf{80.78}
        & 45.32 & 50.00 & 48.98 & 50.65 & \cellcolor{gray!20}\textbf{81.55} \\
        \bottomrule
    \end{tabular}
    \end{adjustbox}
\end{table*}

\subsection{Robustness Against SO(3) Rotations}
Table~\ref{tab:comparison1} reveals a fundamental performance dichotomy under 3D rotational attacks. While RSWP consistently maintains a high Panoramic Bit Accuracy (PBA) across all evaluated angles, baseline methods exhibit a catastrophic performance drop, with accuracies hovering around 50\%, which indicates a near-random guessing. This discrepancy stems from the fundamental limitations of standard 2D CNNs, which rely on translational equivariance defined on a fixed pixel lattice. Qualitative visualizations of $SO(3)$ viewing rotations are provided in Section~S4 of the supplementary material.

In contrast, RSWP effectively mitigates these geometric distortions through its $\mathcal{I}_{60}$-equivariant architecture. For rotations that coincide with elements of the icosahedral symmetry group (e.g., $(0^\circ, 0^\circ, 180^\circ)$ and $(0^\circ, 72^\circ, 0^\circ)$), RSWP achieves near-perfect recovery as feature responses are guaranteed to transform via structured permutations rather than arbitrary distortions.
For general continuous rotations that do not align with discrete group elements (e.g.,$(20^\circ, 40^\circ, 15^\circ)$), RSWP still maintains a high accuracy of above 83.37\%. This robustness can be attributed to the \textit{Multi-Orientation Projector Bank}, which aligns the input with the nearest discrete canonical frame, bounding the residual geometric error to within $\le 5^\circ$. This ``align-and-correct" mechanism ensures that arbitrary continuous rotations are effectively mapped into the model's equivariant comfort zone for reliable recovery.

Additional evaluations of ViDec's robustness to $SO(2)$ viewport rotations are provided in Section~S1 of the supplementary material.

\subsection{Robustness Against Traditional and Combined Attacks}

To rigorously evaluate the practical applicability of RSWP, we extend the assessment beyond pure rotations to include traditional distortions and compound attacks. The quantitative results in Table~\ref{tab:comparison_attacks} demonstrate the robustness of our proposed framework.

The scaling, cropping, and local editing results and their detailed explanations are provided in Section~S6 of the supplementary material.

As shown in Table \ref{tab:comparison_attacks}, RSWP also exhibits high performance under common signal processing attacks. Notably, except for JND and RSWP, all other methods are designed for 2D images and fail to detect viewports effectively. To further evaluate stability under realistic conditions, we consider compound attacks that combine $SO(3)$ rotations with JPEG compression ($Q=80$), Cropout ($20\%$), and Gaussian noise ($\sigma=0.03$). As shown in the bottom half of Table~\ref{tab:comparison_attacks}, baseline methods degrade to near-random guessing under these compound distortions, whereas RSWP maintains high robustness, confirming the effectiveness of our equivariant design in complex real-world scenarios.

Evaluations on different panorama resolutions are provided in Section~S2 of the supplementary material.

The effects of payload size and resolution parameter $\mathcal{S}$ are analyzed in Section~S3 of the supplementary material.

\begin{table*}[t]
\centering
        \caption{Comprehensive ablation study on the key components of RSWP. ``Group Rot. PBA" denotes averaged PBA under rotations from Icosahedral group $\mathcal{I}_{60}$, and ``Arb. Rot. PBA" is the averaged PBA under arbitrary continuous $SO(3)$ rotations.}
\label{tab:ablation_final}
\setlength{\tabcolsep}{3.5pt}
\begin{adjustbox}{max width=\textwidth}
\begin{tabular}{l|ccccc|ccccc}
\toprule
\multirow{2}{*}{\textbf{Model Variant}} & \multicolumn{5}{c|}{\textbf{Component Configuration}} & \multicolumn{5}{c}{\textbf{Performance Metrics}} \\
\cline{2-11}
 & \textbf{Proj. Domain} & \textbf{Net. Type} & \textbf{Sym. Pad} & \textbf{Sorting} & \textbf{Bank} & \textbf{PSNR (dB)} & \textbf{PBA (\%)$\uparrow$} & \textbf{VBA (\%)$\uparrow$} & \textbf{Group Rot.PBA (\%)$\uparrow$} & \textbf{Arb. Rot.PBA (\%)$\uparrow$} \\
\midrule
\multicolumn{11}{l}{\textit{1. Geometric Domain Analysis}} \\
A1 (Baseline) & ERP & CNN & - & - & - & 40.14 & 98.76 & 49.03 & 49.55 & 49.79 \\
A2 (Cubemap) & Cube & CNN & - & - & - & 42.42 & 99.63 & 77.94 & 52.12 & 50.07 \\
A3 (Icosahedral) & Icosa & CNN & - & - & - & 45.47 & 99.70 & 99.01 & 53.80 & 51.25 \\
\midrule
\multicolumn{11}{l}{\textit{2. Equivariance \& Theoretical Constraints}} \\
B1 (No Padding) & Icosa & Eq-CNN & $\times$ & \checkmark & - & 45.80 & 100.00 & 99.73 & 76.33 & 62.15 \\
B2 (No Sorting) & Icosa & Eq-CNN & \checkmark & $\times$ & - &45.86 & 100.00 & 99.91 & 69.17 & 52.49 \\
B3 (Discrete Only) & Icosa & Eq-CNN & \checkmark & \checkmark & $\times$ & 45.86 & 100.00 & 99.91 & 99.70 & 68.50 \\
\midrule
\multicolumn{11}{l}{\textit{3. Full System}} \\
\textbf{RSWP (Ours)} & \textbf{Icosa} & \textbf{Eq-CNN} & \checkmark & \checkmark & \checkmark & \textbf{45.86} & \textbf{100.00} & \textbf{99.91} & \textbf{99.70} & \textbf{88.01} \\
\bottomrule
\end{tabular}
\end{adjustbox}
\end{table*}

\subsection{Ablation Study}
To validate the contribution of each component in RSWP, we conduct a hierarchical ablation study in Table \ref{tab:ablation_final}.

\textbf{1) Impact of Geometric Projection.}
The comparison between Models A1 through A3 isolates the influence of the geometric domain. The baseline Model A1, which applies standard CNNs directly on the ERP plane, fails to handle the inherent spherical distortions, resulting in low accuracies across both viewports and rotated inputs. Transitioning to the icosahedral domain in Model A3 significantly alleviates local geometric distortion, boosting the VBA to near perfection. However, its Group Rot. PBA remains poor, which is comparable to random guessing. This result provides empirical evidence that geometric projection validates viewport extraction. However, while the icosahedral structure simplifies local geometry, it does not automatically confer the global symmetry properties required to withstand 3D rotations without a specialized equivariant network architecture.

\textbf{2) Necessity of Theoretical Constraints.}
We next investigate the critical components required to satisfy strict mathematical equivariance by analyzing Models B1 through B3. Unlike Model A3 which uses a standard CNN, the B-series models adopt the $E(2)$-equivariant backbone (Eq-CNN) as the structural foundation. The results demonstrate that satisfying the group properties of $\mathcal{I}_{60}$ requires rigorous handling of both boundary conditions and feature permutation. Specifically, even with an Eq-CNN backbone, removing the symmetric padding (Model B1) causes the Group Rot. PBA to drop significantly, indicating that boundary leakage can severely impair recovery under rotational transformations. Furthermore, the omission of the permutation-invariant sorting module (Model B2) leads to a complete collapse of the Group Rot. PBA. This confirms that while the network may extract features equivariantly on a per-facet basis, the global synchronization of these features via sorting is a strict prerequisite for correctly decoding the binary sequence. Model B3, which incorporates all these constraints, achieves a near-perfect Group Rot. PBA, validating the theoretical correctness of our discrete equivariant design.

\textbf{3) Bridging Discrete and Continuous Rotations.}
Finally, the comparison between Model B3 and the full RSWP system highlights the necessity of the Multi-Orientation Projector Bank for practical application. Although Model B3 is theoretically sound for discrete rotations, its robustness degrades significantly  when facing arbitrary continuous rotations that fall outside the discrete $\mathcal{I}_{60}$ group. By integrating the Projector Bank, RSWP effectively bridges this gap through mapping arbitrary continuous inputs to the nearest discrete canonical frame. This mechanism ``quantizes" the continuous rotation space, raising the Arb. Rot. PBA while maintaining the high fidelity and theoretical integrity of the system, thereby ensuring robust protection in real-world scenarios.

\section{Conclusion}
In this paper, we present RSWP, a theoretically grounded framework for panoramic watermarking. Its Icosahedral Decoupling Architecture separates spherical rotations into local $SO(2)$ rotations and facet permutations. The Equivariance-to-Invariance decoder combines equivariant feature extraction and permutation-invariant sorting to synchronize watermark recovery, while the viewport decoder enables direct, blind extraction without geometric calibration. The Multi-Orientation Projector Bank bridges discrete icosahedral invariance and robustness to continuous rotations. Experiments demonstrate superior robustness against spherical rotations and viewport transformations while preserving high visual fidelity.

{
    \small
    \bibliographystyle{IEEEtran}

    \bibliography{reference}

}

\end{document}